\documentclass[journal]{IEEEtran}

\usepackage{cite}
\usepackage{amsmath,amssymb,amsfonts}
\usepackage{graphicx}
\usepackage{textcomp}
\usepackage{xcolor}
\usepackage{booktabs}
\usepackage{multirow}
\usepackage{array}
\usepackage{url}
\usepackage{bm}
\usepackage{subcaption}
\usepackage{acronym}
\usepackage[compact]{titlesec}
\usepackage{enumitem} % Required for customizing list labels
\graphicspath{{figures/}{figures/methodology/}{figures/results/}}
\newenvironment{revision}{\color{black}}{}

\titlespacing*{\section}{0pt}{0.6ex plus 0.2ex minus 0.2ex}{0.3ex plus 0.1ex}
\titlespacing*{\subsection}{0pt}{0.4ex plus 0.2ex minus 0.1ex}{0.2ex plus 0.1ex}
\titlespacing*{\subsubsection}{0pt}{0.3ex plus 0.1ex minus 0.1ex}{0.15ex plus 0.05ex}

\newcommand{\G}{\mathcal{G}}
\newcommand{\V}{\mathcal{V}}
\newcommand{\E}{\mathcal{E}}

\newcommand{\D}{\mathcal{D}}
\newcommand{\T}{\mathcal{T}}

\begin{document}

% Define GFM acronym directly (single-pass, no aux file required)
\newacro{GFM}[GFM]{graph foundation model}

\title{Scalable Heterogeneous Graph Foundation-Model Workflows for Data-Driven Optimal Power Flow}

\author{Massimiliano Lupo Pasini, Yijiang Li, Kibaek Kim, and Teja Kuruganti
\thanks{M. Lupo Pasini and T. Kuruganti are with the Computational Sciences and Engineering Division, Oak Ridge National Laboratory, 1 Bethel Valley Road, Oak Ridge, TN 37831, USA.}
\thanks{Y. Li and K. Kim are with the Mathematics and Computer Science Division, Argonne National Laboratory, 9700 S. Cass Avenue, Lemont, IL 60439.}
\thanks{Manuscript received Month XX, 2026; revised Month XX, 2026.}
\thanks{M. Lupo Pasini thanks Dr. Vladimir Protopopescu for the advice received in the drafting of the manuscript.}
\thanks{This manuscript has been authored in part by UT-Battelle, LLC, under contract DE-AC05-00OR22725 with the US Department of Energy (DOE). The US government retains and the publisher, by accepting the article for publication, acknowledges that the US government retains a nonexclusive, paid-up, irrevocable, worldwide license to publish or reproduce the published form of this manuscript, or allow others to do so, for US government purposes. DOE will provide public access to these results of federally sponsored research in accordance with the DOE Public Access Plan (\protect\url{http://energy.gov/downloads/doe-public-access-plan}). This research used resources of the Oak Ridge Leadership Computing Facility at Oak Ridge National Laboratory, which is supported by the Office of Science of the U.S. Department of Energy under Contract No. DE-AC05-00OR22725.  }}

\maketitle

\begin{abstract}
\begin{revision}
Fast and reliable optimal power flow (OPF) approximation is important for power system operation, yet heterogeneous OPF graph models are often evaluated either with architecture-specific implementations or on limited training corpora. This paper presents a large-scale heterogeneous graph-learning workflow, built on HydraGNN, for data-driven OPF surrogate modeling and graph foundation-model (GFM) development. The workflow preserves buses, generators, loads, shunts, alternating-current (AC) lines, transformers, and device-to-bus relations and provides a common implementation for distributed preprocessing, multi-graphics-processing-unit (GPU) training, hyperparameter optimization (HPO), and downstream adaptation. Using approximately three million heterogeneous graphs from ten Power Grid Library for Benchmarking AC Optimal Power Flow Algorithms (PGLib-OPF) cases spanning 14 to 13,659 buses, we compare six heterogeneous graph neural network (GNN) families under a common HPO campaign on Frontier, resulting in two compact HeteroSAGE and HeteroHEAT models ($\sim$1.6-1.7 million parameters) that achieve the lowest observed validation MSE. Strong and weak scaling using up to 1,024 Frontier nodes show that 256 nodes (corresponding to 1,024 AMD MI250X GPUs) provides the best time-to-solution balance for the measured workload. Two downstream tasks evaluate task adaptation on the IEEE 118-bus case: a classification task to distinguish nominal operating conditions from artificially generated extreme-overload conditions and $N{-}1$ single-component-outage regression to predict bus voltage magnitude and phase angle. Partial fine-tuning provides the strongest balance between predictive performance and computational requirements.
\end{revision}
\end{abstract}


% ============================================================
\section{Introduction}
% ============================================================

\IEEEPARstart{O}{ptimal} power flow (OPF) is a fundamental computational primitive for power-grid operation, planning, and decision support~\cite{cain2012opfhistory,frank2012opfsurvey}. Given a grid topology, component parameters, generator capabilities, and load demands, the alternating current OPF (AC-OPF) problem seeks a feasible operating point that minimizes generation cost while satisfying both voltage, current, and power flow (PF) demands, and equipment constraints. The solution contains bus voltage magnitudes and voltage angles, from which additional electrical quantities such as branch flows, currents, and losses can be derived.

Modern smart grids increasingly involve heterogeneous devices, frequent changes in operating conditions, distributed and controllable resources, topology variations, and the need for fast decision support. These trends motivate learning-based OPF surrogates that can approximate OPF solutions with \begin{revision}fast predictions and  may eventually support rapid screening of large volumes of grid scenarios\end{revision}. 

%Recent reviews summarize the growing range of data-driven OPF surrogate formulations~\cite{mohammadi2024opfreview}. Prior work has explored imitation-learning graph neural networks (GNNs) for OPF approximation~\cite{owerko2019opfgnn}, topology-aware GNNs for feasible and adaptive AC-OPF prediction~\cite{liu2022topologyaware}, neural architectures derived from iterative AC-OPF algorithms~\cite{guo2023dagnn}, GNN-assisted solver initialization~\cite{deihim2024initialestimate}, and physics-informed neural approaches that embed AC-OPF constraints into training~\cite{nellikkath2022physics,fioretto2020predicting}. Recent studies have further introduced heterogeneous graph representations and attention mechanisms for AC-OPF~\cite{ghamizi2024opfhgnn,trigui2025attentionopf,wen2026localglobalhgnn}, while LUMINA and LUMINA-Bench have advanced foundation-model-style AC-OPF learning through topology transfer, feasibility-aware objectives, standardized benchmarks, zero-shot generalization, and fine-tuning protocols~\cite{li2026lumina,jin2026luminabench,memon2026systematic}. 

\begin{revision}
% ============================================================
\subsection{State-of-the-Art on Data-Driven AC-OPF Surrogates}
\label{sec:related_work}
% ============================================================

Recent reviews document the growing use of data-driven surrogate models for
OPF%, including fully connected neural networks, GNNs, physics-informed models, and hybrid approaches that combine learned predictions with conventional optimization algorithms
\cite{mohammadi2024opfreview}. State-of-the-art (SOTA) work in this field can
be organized into four areas: (1) graph-based AC-OPF and PF surrogates, (2) feasibility-aware and physics-informed OPF learning, (3) heterogeneous graph representations for power systems, and (4) grid foundation models, benchmarks, and open frameworks.

\subsubsection{Graph-Based AC-OPF and PF Surrogates}
\label{subsec:gnn_opf_lit}
Power networks have a natural graph structure in which buses and electrical
devices are connected through transmission lines and transformers. GNNs can exploit this structure by sharing
information locally according to the physical connectivity of the network.
Early graph-based OPF work formulated OPF approximation as an
imitation-learning problem over graph-structured grid data
\cite{owerko2019opfgnn}. Subsequent topology-aware approaches used the
structural locality of AC-OPF quantities to improve prediction and adaptation
under changes in network configuration \cite{liu2022topologyaware}.

Other studies have used graph-based models to emulate or accelerate specific
parts of the numerical solution process. Directed-acyclic neural
architectures have been derived from iterative AC-OPF solution mappings
\cite{guo2023dagnn}, while GNN-generated initial estimates have been used to
accelerate the convergence of interior-point OPF solvers
\cite{deihim2024initialestimate}. These methods show the value of
topology-aware representations, but they target a particular
architecture, solver stage, or limited collection of grid cases.

Closely related work also considers AC power flow approximation rather than
optimization. PowerFlowNet uses message-passing GNNs to approximate AC
power flow solutions on networks ranging from the IEEE 14-bus system to the
French 6470rte network \cite{lin2024powerflownet}. Because PowerFlowNet
predicts solutions of the power flow equations without optimizing an
operating-cost objective, it is relevant evidence for graph-based grid
surrogates but is not a direct AC-OPF baseline. %Topology-aware neural networks have also been investigated for rapid contingency analysis across $N{-}1$ and higher-order outage scenarios \cite{nakiganda2023contingency}. This line of work is particularly relevant to downstream tasks involving network reconfiguration, although its primary objective is contingency screening or power flow evaluation rather than multi-case AC-OPF pretraining.

\subsubsection{Feasibility-Aware and Physics-Informed OPF Learning}
\label{subsec:physics_informed_opf_lit}

A low prediction error does not necessarily guarantee that a learned OPF solution
satisfies the nonlinear power flow equations and engineering operating
constraints. Physics-informed AC-OPF models address this limitation by
incorporating power flow residuals and constraint violations into the
training objective \cite{nellikkath2022physics}. Lagrangian-dual training
has similarly been used to balance prediction accuracy against physical and
engineering constraints \cite{fioretto2020predicting}. Verification-based
methods provide a complementary perspective by computing worst-case
guarantees on constraint violations, prediction error, and suboptimality for
learned OPF mappings \cite{venzke2020worstcase}.
These studies show that feasibility-aware learning can improve physical
consistency, but they also expose a typical challenge of multi-objective optimization problems:
the data-fitting term and the constraint terms may favor different parameter
updates during training. Their relative influence depends on the penalty
weights, the scaling of the individual residuals, the frequency with which
constraints become active, and the grid systems represented in the training
corpus. 

\subsubsection{Heterogeneous Graph Representations for Power Systems}
\label{subsec:heterogeneous_opf_lit}

Power systems contain distinct physical component types, including buses,
generators, loads, shunts, transmission lines, and transformers.
Homogeneous graph representations typically treat different devices and relations uniformly. This
simplification obscures differences in component features, relation
semantics, and edge attributes. Heterogeneous graph representations instead
retain explicit node and edge types, allowing the message-passing operation
to depend on the source component, destination component, and physical
relation.

OPF-HGNN introduced a heterogeneous GNN formulation for AC-OPF
generalization with differentiable grid-constraint penalties
\cite{ghamizi2024opfhgnn}. Related studies have examined heterogeneous
attention mechanisms for AC-OPF \cite{trigui2025attentionopf} and
architectures that combine local and global heterogeneous message passing
\cite{wen2026localglobalhgnn}. These models establish the relevance of typed power system graphs but do not provide the same combination of multi-architecture model selection, reusable distributed preprocessing, leadership-scale training, and downstream adaptation.

%The heterogeneous HydraGNN models evaluated here build on established message-passing and attention operators, including GraphSAGE \cite{hamilton2017graphsage}, GAT \cite{velickovic2018gat}, GATv2 \cite{brody2022gatv2}, PNA\cite{corso2020pna}, HGT \cite{hu2020hgt}, and HEAT \cite{mo2021heat}. The contribution of the present work is not the introduction of these underlying operators. It is their integration into a common heterogeneous OPF workflow in which graph construction, distributed data access, training, hyperparameter optimization, and downstream adaptation are held consistent across architecture families.

\subsubsection{Grid Foundation Models, Benchmarks, and Open Frameworks}
\label{subsec:grid_foundation_models}

Recent work has moved beyond topology-specific surrogates toward grid
foundation models that learn reusable representations across multiple
systems, operating conditions, and downstream tasks. LUMINA formulates
AC-OPF surrogate learning as a foundation-model problem and studies design
principles related to topology transfer, feasibility-aware learning, and
reliability under distribution shift \cite{li2026lumina}. LUMINA-Bench
provides standardized protocols for single-topology training,
multi-topology pretraining, held-out-topology generalization, zero-shot
evaluation, and fine-tuning, with metrics for both prediction error and
constraint violations \cite{jin2026luminabench}. Related LUMINA work extends
this generalization perspective across coupled grid-optimization problems,
including unit commitment (UC), security-constrained unit commitment
(SCUC), AC-OPF, and integrated UC--AC-OPF formulations
\cite{memon2026systematic}. However, existing work has shown moderate scalability with distributed training on selected sets of topologies that do not exceed 2,000 buses. 

Microsoft GridSFM provides a complementary domain-specific model and open
release for multi-topology AC-OPF inference \cite{gridsfm2026}. Its open
model documentation reports approximately 540,000 scenarios spanning 54
base topologies, together with physics-aware output processing and
analytical reconstruction of branch quantities. GridSFM is therefore a relevant comparison for multi-topology AC-OPF learning, although its training corpus, model architecture, prediction targets, and evaluation protocol differ from those used in this work.

At a broader level, IBM GridFM presents a research agenda and implementation
roadmap for foundation models trained across heterogeneous electric-grid
data and topologies \cite{hamann2024gridfm}. OpenGridFM, originating from
Hydro-Qu\'ebec and IBM and now hosted by LF Energy, provides an open-source
community framework for developing and evaluating grid foundation models
\cite{opengridfm2026}. IBM GridFM is primarily a research vision, whereas OpenGridFM is an enabling software and community framework; neither provides a directly comparable AC-OPF benchmark under the data, targets, and splits used in the present study.

%Among these efforts, OPF-HGNN, LUMINA, LUMINA-Bench, and GridSFM provide the most direct methodological context for the present work because they address heterogeneous, transferable, or multi-topology AC-OPF learning. PowerFlowNet is an adjacent power flow surrogate, while IBM GridFM and OpenGridFM provide broader conceptual and software context. Table~\ref{tab:novelty_comparison} summarizes these distinctions and positions the HydraGNN OPF workflow relative to the corresponding methods, benchmarks, and open frameworks.

\subsection{Outstanding Challenges}
SOTA efforts show the potential of graph-based OPF learning, but provide limited answers to the following five questions: (i) which model configurations achieve the best predictive validation loss under a common benchmark; (ii) whether the complete workflow can execute model selection while training each model over the entire data corpus comprising millions of heterogeneous graphs on leadership-class resources; (iii) how optimization robustness varies across heterogeneous GNN families; (iv) how the best performing GNN architectures scale under fixed workload and fixed local batch protocols on leadership-class computing facilities; and (v) how different fine-tuning strategies (which differ in the set of model parameters that are frozen and adapted) compare in terms of predictive performance on downstream tasks of different type, namely classification and regression. 

\end{revision}

\subsection{Our contributions}

%\begin{revision}These efforts show the potential of graph-based OPF learning, but provide limited evidence on the capabilities required to train and evaluate heterogeneous OPF graph models on multi-topology datasets containing millions of samples using leadership-class computing resources.\end{revision} Existing benchmark-oriented studies provide controlled comparisons across architectures, losses, transfer settings, and feasibility metrics, but full-corpus distributed preprocessing, large-scale multi-architecture HPO, and downstream fine-tuning introduce additional challenges in memory footprint, communication cost, data loading, and training stability.

\begin{revision}
%The underlying GraphSAGE, graph-attention (GAT), heterogeneous graph transformer (HGT), heterogeneous edge-enhanced graph attention (HEAT), and principal neighborhood aggregation (PNA) operators are established methods; the contribution is their integration behind a common heterogeneous interface that shares graph construction, distributed preprocessing, data access, training, HPO, and adaptation for fine-tuning. This distinction makes the architecture itself an experimental variable without rebuilding the data and execution pipeline for each model family.
Our work addresses these challenges through the following scientific and computer science contributions:
\begin{enumerate}[label=(\roman*)]
    \item an extension of HydraGNN\footnote{\url{https://github.com/ORNL/HydraGNN}}~\cite{hydragnn2021} with a configurable heterogeneous graph-learning workflow for OPF surrogate modeling and \ac{GFM} pretraining,  enabling to benchmark six families of GNNs for heterogeneous graphs within a unified framework, namely GraphSage, principal neighborhood aggregation (PNA), graph-attention (GAT), relational graph attention (RGATs), heterogeneous graph transformer (HGT), and heterogeneous edge-enhanced graph attention (HEAT); 
    \item a large scale HPO campaign on 512 nodes of Frontier that compares six heterogeneous GNN families over approximately three million graphs from ten power system cases spanning 14 to 13,659 buses, the HPO campaign identified compact GNN models with the lowest validation losses among completed HPO trials;
    \item {\color{black}a systematic characterization of the predictive performance, hyperparameter sensitivity, and optimization robustness of six heterogeneous GNN architecture families for AC-OPF surrogate modeling, identifying compact GNN architectures as the strongest-performing candidates under the evaluated HPO protocol;}
    \item strong and weak scaling measurements on 1--1,024 Frontier nodes, identifying approximately 256 nodes as the practical saturation point for the measured workload; and
    \item developing multiple fine-tuning strategies within the HydraGNN framework and testing them on fine-tuning tasks using the grid topology from IEEE case118, comprising a classification task to reliably and efficiently distinguish nominal operating conditions from artificially generated extreme-overload conditions and an $N{-}1$ single-component-outage regression task. 
\end{enumerate}

{\textcolor{black}{
These contributions result in a unified, scalable computational workflow that makes systematic, full-corpus exploration of heterogeneous GNN architectures practical within a reasonable experimental timeframe.}
}

Table~\ref{tab:novelty_comparison} positions our HydraGNN-based workflow with respect to prior GNN-based OPF surrogate modeling and foundation model approaches. IBM GridFM and OpenGridFM are
not included in Table~\ref{tab:novelty_comparison}, because IBM GridFM
primarily presents a research vision and implementation roadmap, whereas
OpenGridFM is an enabling software and community framework rather than a
specific trained AC-OPF model with reported data, HPO, transfer, and
computational-scaling results that can be compared consistently with the
empirical approaches summarized in the table.

\begin{table*}[t]
\centering
\scriptsize
\setlength{\tabcolsep}{2.2pt}
\renewcommand{\arraystretch}{1.15}
\caption{Capability comparison between the HydraGNN OPF workflow
and closely related graph-based power-flow and AC-OPF approaches.
``NR'' means that the corresponding capability or computational scale was
not reported in the cited publication or public release; it does not imply
that the capability cannot be implemented. Dataset sizes refer to the
reported training corpus or, when explicitly indicated, the complete
released dataset.}
\label{tab:novelty_comparison}
\begin{tabular}{
    p{0.145\textwidth}
    p{0.175\textwidth}
    p{0.175\textwidth}
    p{0.155\textwidth}
    p{0.155\textwidth}
    p{0.155\textwidth}
}
\toprule
\textbf{Capability}
&
\textbf{HydraGNN OPF workflow (this work)}
&
\textbf{LUMINA / LUMINA-Bench
\cite{li2026lumina,jin2026luminabench}}
&
\textbf{Microsoft GridSFM
\cite{gridsfm2026}}
&
\textbf{OPF-HGNN
\cite{ghamizi2024opfhgnn}}
&
\textbf{PowerFlowNet
\cite{lin2024powerflownet}}
\\
\midrule

\textbf{Learning task}
&
AC-OPF bus-state regression, followed by feasibility classification and
$N{-}1$ contingency regression
&
AC-OPF regression, feasibility-aware learning, topology transfer, and
fine-tuning
&
AC-OPF state and dispatch prediction, feasibility classification, and
solver warm starting
&
AC-OPF generator dispatch prediction
&
AC power-flow state approximation; no OPF objective
\\

\textbf{Graph representation}
&
Heterogeneous graph with buses, generators, loads, shunts, AC lines,
transformers, and typed device-to-bus relations
&
Homogeneous and heterogeneous graph representations
&
Heterogeneous grid embedding with transformer-style processing and
multiple output heads
&
Heterogeneous representation of buses, generators, loads, external-grid
components, and lines
&
Homogeneous bus--branch graph
\\

\textbf{Supported GNN operators}
&
Six heterogeneous families---HeteroSAGE, HeteroGAT, HeteroRGAT,
HeteroHEAT, HeteroHGT, and HeteroPNA---implemented through a shared
HydraGNN heterogeneous-model interface
&
Eight homogeneous and heterogeneous backbones---GCN, GAT, GIN, graph
Transformer, RGAT, HeteroGNN, HGT, and HEAT
&
One domain-specific transformer architecture with topological diffusion
&
One heterogeneous GraphSAGE architecture; homogeneous GraphSAGE and
fully connected networks are used as baselines
&
One specialized message-passing architecture, PowerFlowConv\\

\textbf{Training-data scale}
&
Approximately $3$ million graphs from ten PGLib-OPF cases
&
Uses OPFData, whose complete corpus contains approximately $3$ million
instances over ten cases; controlled multi-topology experiments primarily
use selected subsets
&
Approximately $540{,}000$ scenarios spanning 54 base topologies in the
open release
&
$8{,}000$ training and $2{,}000$ validation graphs for each of three
systems: 9-, 14-, and 30-bus
&
$180{,}000$ released scenarios: $100{,}000$, $50{,}000$, and $30{,}000$
for the 14-, 118-, and 6470-bus systems, respectively
\\

\textbf{Grid-size range}
&
14--13,659 buses
&
30--2,000 buses
&
500--4,661 buses
&
9--30 buses
&
14--6,470 buses
\\

\textbf{Hyperparameter optimization}
&
DeepHyper centralized Bayesian HPO across six architecture
families
&
Bayesian HPO is sequential and not distributed
&
Systematic HPO procedure and search scale are NR
&
Fixed architecture and training settings are reported; systematic HPO
is NR
&
Architectural and loss ablations are reported; systematic HPO is NR
\\

\textbf{Distributed training and HPC scale}
&
HPO uses 512 Frontier nodes per architecture campaign, with 32 nodes
and 256 GPUs per trial. Strong and weak scaling are measured through
1,024 Frontier nodes
&
Uses ALCF and NERSC resources, but the number of compute nodes or GPUs,
distributed-training configuration, and scaling efficiency are NR
&
Distributed-training configuration, maximum node count, and scaling
efficiency are NR
&
Distributed training and multi-node scaling are NR
&
Experiments use one NVIDIA RTX 3060 Ti GPU; distributed training is NR
\\

\textbf{Scalable data pipeline}
&
Distributed conversion to 129 reusable HDF5 shards and distributed
multi-GPU loading of the full multi-case corpus
&
Reusable data-processing pipeline is provided; distributed preprocessing
and large-scale I/O measurements are NR
&
Open data-generation and inference pipeline; distributed training-data
I/O performance is NR
&
PandaPower-based mutation and simulation pipeline; distributed
preprocessing is NR
&
Open synthetic-data generator and datasets; distributed preprocessing
is NR
\\

\textbf{Physics and feasibility}
&
Voltage-magnitude, angle-difference, and approximate thermal-limit
penalties
&
MSE, augmented-Lagrangian, and violation-based Lagrangian objectives;
explicit power-balance and operating-limit metrics
&
Predicts feasibility and reconstructs branch quantities for
physics-oriented evaluation
&
Differentiable constraint penalties are included in the node-level loss
&
Power-balance and mixed supervised/physical losses are evaluated
\\

\textbf{Transfer evaluation}
&
Fine-tuning is evaluated for two downstream tasks on case118; held-out
topology transfer is not claimed
&
Single- and multi-topology pretraining, zero-shot evaluation,
held-out-topology transfer, and fine-tuning
&
Multi-topology pretraining, out-of-distribution evaluation, and
few-shot adaptation to new grids
&
Generalization across load and generation-cost perturbations on three
small systems
&
Cross-system and mixed-system experiments for power-flow prediction
\\

\textbf{Capability distinguishing this work}
&
Combines a common multi-architecture interface with full-corpus
distributed preprocessing, concurrent distributed Bayesian HPO, and
measured strong and weak scaling on up to 1,024 Frontier nodes
&
Provides a common multi-architecture framework. Bayesian HPO is performed sequentially with
Optuna, and multi-node computational scaling is not evaluated
&
Provides greater topology diversity, but does not report
multi-architecture HPO or multi-node training scalability
&
Provides heterogeneous and feasibility-aware learning at comparatively
small data and grid scales
&
Addresses power flow rather than OPF and does not provide a
heterogeneous multi-architecture or distributed workflow
\\

\bottomrule
\end{tabular}
\end{table*}

\end{revision}

% ============================================================
\section{Heterogeneous Graph Learning for Optimal Power Flow}
\label{sec:heterogeneous_graph_learning}
% ============================================================

% ============================================================
\subsection{Heterogeneous Graph Representation of Power Grids}
\label{sec:heterogeneous_opf_graph}
% ============================================================

\subsubsection{Node Types and Node Features}
\label{subsec:node_features}

Each OPF instance is represented as a heterogeneous graph
\begin{equation}
    \G = (\V, \E),
\end{equation}
where the node set is decomposed into four typed subsets:
\begin{equation}
    \V = \V_{\mathrm{bus}} \cup \V_{\mathrm{gen}} \cup \V_{\mathrm{load}} \cup \V_{\mathrm{shunt}}.
\end{equation}

Table~\ref{tab:opf_node_features} summarizes the node types and their input-feature dimensions. The bus node is the target node type for the primary supervised task, while generators, loads, and shunts participate in message passing to inform bus-level predictions.

\begin{revision}
\begin{table}[t]
\centering
\scriptsize
\setlength{\tabcolsep}{3pt}
\caption{Node inputs and prediction targets, following the OPFData schema~\cite{lovett2024opfdata}.}
\label{tab:opf_node_features}
\renewcommand{\arraystretch}{1.1}
\begin{tabular}{p{1.15cm}cp{4.95cm}}
\toprule
\textbf{Type} & \textbf{Dim.} & \textbf{Input features and role} \\
\midrule
bus & 4 & Voltage lower/upper bounds, base voltage, and bus type; structural operating context, not the predicted voltage state \\
generator & 11 & Active/reactive dispatch and limits, status/base information, and cost coefficients supplied by OPFData \\
load & 2 & Active and reactive demand $(P_d,Q_d)$ \\
shunt & 2 & Conductance and susceptance $(G_s,B_s)$ \\
\midrule
bus target & 2 & Voltage angle $V_a$ (rad) and magnitude $V_m$ in per-unit (p.u.) notation \\
\bottomrule
\end{tabular}
\end{table}
\end{revision}

Bus features encode electrical state constraints and bus metadata, such as voltage limits, base voltage, and bus type. Generator features encode capacity limits, active and reactive power information, cost coefficients, current dispatch setpoints, and operating status. Load features encode active and reactive demand. Shunt features encode conductance and susceptance.

\subsubsection{Edge Types and Edge Features}
\label{subsec:edge_features}

The edge set $\E$ is the union of both physical electrical edges and structural equipment-to-bus links:
\begin{equation}
    \E = \E_{\mathrm{ac}} \cup \E_{\mathrm{tr}} \cup \E_{\mathrm{gen}} \cup \E_{\mathrm{load}} \cup \E_{\mathrm{shunt}}.
\end{equation}

Each edge relation is represented as a triplet consisting of source node type, edge type, and target node type. Table~\ref{tab:opf_edge_relations} lists the relations used in the heterogeneous OPF graph.

\begin{table}[t]
\centering
\small
\setlength{\tabcolsep}{4pt}
\caption{Typed edge relations in the heterogeneous OPF graph.}
\label{tab:opf_edge_relations}
\renewcommand{\arraystretch}{1.15}
\begin{tabular}{llp{1.6cm}p{3.1cm}}
\toprule
\textbf{Source} & \textbf{Relation} & \textbf{Target} & \textbf{Physical Meaning} \\
\midrule
bus       & ac\_line        & bus       & AC line between buses \\
bus       & transformer     & bus       & Transformer between buses \\
generator & generator\_link & bus       & Generator attached to bus \\
bus       & generator\_link & generator & Reverse generator link \\
load      & load\_link      & bus       & Load attached to bus \\
bus       & load\_link      & load      & Reverse load link \\
shunt     & shunt\_link     & bus       & Shunt attached to bus \\
bus       & shunt\_link     & shunt     & Reverse shunt link \\
\bottomrule
\end{tabular}
\end{table}

Only the two physical edge types carry continuous edge attributes, as summarized in Table~\ref{tab:opf_edge_features}. The structural links carry no continuous features but are included in both directions to enable bidirectional information flow between buses and attached equipment.

\begin{revision}
\begin{table}[t]
\centering
\scriptsize
\setlength{\tabcolsep}{3pt}
\caption{Relation-dependent edge attributes in the OPFData representation~\cite{lovett2024opfdata}.}
\label{tab:opf_edge_features}
\renewcommand{\arraystretch}{1.1}
\begin{tabular}{p{1.45cm}cp{4.65cm}}
\toprule
\textbf{Relation} & \textbf{Dim.} & \textbf{Attributes} \\
\midrule
AC line & 9 & Angle-difference limits, series resistance/reactance, terminal shunt susceptances, and three thermal ratings \\
transformer & 11 & Angle limits, series parameters, terminal shunt terms, thermal ratings, tap ratio, and phase shift \\
device--bus links & 0 & Structural connectivity only; reverse relations are also included \\
\bottomrule
\end{tabular}
\end{table}
\end{revision}

AC-line features describe electrical characteristics of a transmission line under a pi-equivalent model, including minimum and maximum angle difference, series resistance, series reactance, shunt susceptance terms, and thermal ratings. Transformer features include angle limits, series resistance, leakage reactance, off-nominal turns ratio, thermal ratings, tap setting, and phase shift. This distinction motivates heterogeneous architectures that can process relation-specific edge attributes with different dimensionalities.

% ============================================================
\section{Fully Supervised OPF Learning Task}
\label{sec:supervised_opf_task}
% ============================================================

\subsection{OPF Surrogate Learning as a Smart Grid Data-Analytics Task}
\label{subsec:smart_grid_surrogate}

In smart grid applications, OPF surrogates are useful when many operating scenarios, topology perturbations, or control decisions must be evaluated quickly. Examples include real-time screening, contingency analysis, approximate dispatch, solver warm-starting, and rapid evaluation of operating-condition changes. In this paper, the learned model approximates the OPF solution map
\begin{equation}
    f_\star : \mathcal{X} \rightarrow \mathcal{Y},
\end{equation}
where $\mathcal{X}$ denotes grid topology, component parameters, and operating context, and $\mathcal{Y}$ denotes OPF solution quantities. The primary task considered in this paper is bus-level prediction of voltage angles and voltage magnitudes.

\subsection{Input Features}
\label{subsec:input_features}

The model input for one OPF sample is a heterogeneous graph
\begin{equation}
    \G = \left( \{\mathbf{X}^{\tau}\}_{\tau \in \T_v},
    \{\mathbf{E}^{r}, \mathbf{A}^{r}\}_{r \in \T_e}, \mathbf{g} \right),
\end{equation}
where $\mathbf{X}^{\tau}$ is the node-feature matrix for node type $\tau$, $\mathbf{E}^{r}$ is the edge-index tensor for relation $r$, $\mathbf{A}^{r}$ is the edge-attribute matrix when available, and $\mathbf{g}$ denotes optional graph-level context features.

The input features include bus, generator, load, and shunt node features, AC-line and transformer edge attributes, and structural edge connectivity. For the primary bus-level prediction task, all node and edge types participate in message passing, but the decoder is applied only to bus-node embeddings.

\begin{revision}

\subsection{Output Targets and Bus-Node Prediction Head}
\label{subsec:output_targets}

Let
\begin{equation}
    \mathbf{h}_i^{(L)}
    =
    f_{\Theta_{\mathrm{enc}}}^{\mathrm{enc}}(\G)_i
\end{equation}
denote the final embedding of bus node
$i\in\V_{\mathrm{bus}}$ after the last heterogeneous message-passing
layer, with $L$ denoting the total number of message passing layers. The pretraining prediction head is applied independently to the
final embedding of each bus:
\begin{equation}
    \widehat{\mathbf{y}}_i
    =
    f_{\Theta_{\mathrm{head}}}^{\mathrm{bus}}
    \left(\mathbf{h}_i^{(L)}\right)
    =
    \begin{bmatrix}
        \widehat{V}_{a,i} \\
        \widehat{V}_{m,i}
    \end{bmatrix},
    \qquad i\in\V_{\mathrm{bus}},
    \label{eq:bus_prediction_head}
\end{equation}
where $\widehat{V}_{a,i}$ and $\widehat{V}_{m,i}$ are, respectively,
the predicted voltage angle in radians and voltage magnitude in per unit
for bus $i$. Equivalently, the complete encoder--head model can be written
as
\begin{equation}
    f_{\Theta_{\mathrm{enc}},\Theta_{\mathrm{head}}}(\G)_i
    =
    f_{\Theta_{\mathrm{head}}}^{\mathrm{bus}}
    \left(
        f_{\Theta_{\mathrm{enc}}}^{\mathrm{enc}}(\G)_i
    \right).
    \label{eq:encoder_head_composition}
\end{equation}

Generators, loads, shunts, AC lines, transformers, and structural
equipment-to-bus relations contribute to the bus embeddings through
heterogeneous message passing. The prediction head is applied only to
bus-node embeddings because the pretraining task predicts bus-level
quantities. The other node types contribute contextual information to
the bus representations but do not receive supervised output labels
during pretraining.

The supervised target at bus $i$ is
\begin{equation}
    \mathbf{y}_i
    =
    \begin{bmatrix}
        V_{a,i} \\
        V_{m,i}
    \end{bmatrix}.
\end{equation}
The supervised loss is the bus-level mean squared error:
\begin{equation}
    \mathcal{L}_{\mathrm{sup}}
    =
    \frac{1}{|\V_{\mathrm{bus}}|}
    \sum_{i\in\V_{\mathrm{bus}}}
    \left\|
        \widehat{\mathbf{y}}_i-\mathbf{y}_i
    \right\|_2^2.
    \label{eq:supervised_loss}
\end{equation}

\subsection{Feasibility-Aware Loss Functions}
\label{subsec:preliminary_feasibility_losses}

In addition to the supervised objective in
Equation~\eqref{eq:supervised_loss}, HydraGNN allows to add
soft penalties to the training loss intended to discourage violations of selected OPF operating
limits, following the recommendations from recent feasibility-aware OPF
surrogates showing that a low prediction error does not necessarily imply a
physically feasible solution~\cite{nellikkath2022physics,fioretto2020predicting,ghamizi2024opfhgnn}.

For each bus $i\in\mathcal{V}_{\mathrm{bus}}$, let $\widehat{V}_{m,i}$ denote
the predicted voltage magnitude and let
$V_{m,i}^{\min}$ and $V_{m,i}^{\max}$ denote its operating limits. The
voltage-bound penalty is defined using a squared hinge loss:
\begin{equation}
\begin{split}
    \mathcal{L}_{V}
    =
    \frac{1}{|\mathcal{V}_{\mathrm{bus}}|}
    \sum_{i\in\mathcal{V}_{\mathrm{bus}}}
    \Big[
        &\operatorname{ReLU}
        \left(V_{m,i}^{\min}-\widehat{V}_{m,i}\right)^2 \\
        +\,
        &\operatorname{ReLU}
        \left(\widehat{V}_{m,i}-V_{m,i}^{\max}\right)^2
    \Big].
\end{split}
\label{eq:voltage_bound_penalty}
\end{equation}

For each AC line or transformer $(i,j)\in\mathcal{E}_{\mathrm{phys}}$, the
predicted phase-angle difference is
\begin{equation}
    \widehat{\Delta\theta}_{ij}
    =
    \widehat{V}_{a,i}-\widehat{V}_{a,j},
\end{equation}
where $\widehat{V}_{a,i}$ and $\widehat{V}_{a,j}$ are the predicted bus
voltage angles. 
Let $\mathcal{E}_{\mathrm{phys}}$ denote the set of physical branches,
including AC transmission lines and transformers, and let
$|\mathcal{E}_{\mathrm{phys}}|$ denote its cardinality.
Given the branch-specific limits
$\Delta\theta_{ij}^{\min}$ and $\Delta\theta_{ij}^{\max}$, the corresponding
penalty is
\begin{equation}
\begin{split}
    \mathcal{L}_{\theta}
    =
    \frac{1}{|\mathcal{E}_{\mathrm{phys}}|}
    \sum_{(i,j)\in\mathcal{E}_{\mathrm{phys}}}
    \Big[
        &\operatorname{ReLU}
        \left(
        \Delta\theta_{ij}^{\min}
        -
        \widehat{\Delta\theta}_{ij}
        \right)^2 \\
        +\,
        &\operatorname{ReLU}
        \left(
        \widehat{\Delta\theta}_{ij}
        -
        \Delta\theta_{ij}^{\max}
        \right)^2
    \Big].
\end{split}
\label{eq:angle_bound_penalty}
\end{equation}

Predicted terminal
apparent-power flows,
$\widehat{S}_{ij}$ and $\widehat{S}_{ji}$, are reconstructed from the
predicted complex bus voltages and the available branch parameters. For a
branch with thermal rating $S_{ij}^{\max}$, the penalty is
\begin{equation}
\begin{split}
    \mathcal{L}_{S}
    =
    \frac{1}{|\mathcal{E}_{\mathrm{phys}}|}
    \sum_{(i,j)\in\mathcal{E}_{\mathrm{phys}}}
    \Big[
        &\operatorname{ReLU}
        \left(
        |\widehat{S}_{ij}|-S_{ij}^{\max}
        \right)^2 \\
        +\,
        &\operatorname{ReLU}
        \left(
        |\widehat{S}_{ji}|-S_{ij}^{\max}
        \right)^2
    \Big].
\end{split}
\label{eq:thermal_penalty}
\end{equation}
The penalties $\mathcal{L}_{\theta}$ and $\mathcal{L}_{S}$ are averaged over
the physical branches using
$\frac{1}{|\mathcal{E}_{\mathrm{phys}}|}$ as a scaling factor, so that their
magnitudes does not grow directly with graph size.

The three penalty terms have different physical units
and, consequently, very different raw magnitudes: $\mathcal{L}_{V}$ is
measured in squared per-unit voltage, $\mathcal{L}_{\theta}$ in squared
radians, and $\mathcal{L}_{S}$ in squared per-unit apparent power. 
To make the three weights directly comparable, each raw penalty
$\mathcal{L}_k(t)$, $k\in\{V,\theta,S\}$, evaluated at training step $t$, is rescaled by
an exponential moving average (EMA) of its own magnitude before the
corresponding weight is applied. The EMA is defined as
\begin{equation}
    \bar{\mathcal{L}}_k(t) =
    \begin{cases}
        \max\!\big(\mathcal{L}_k(t),\,\varepsilon\big), & t = t_0 \\[4pt]
        \max\!\Big(\beta\, \mathcal{L}_k(t) + (1-\beta)\,\bar{\mathcal{L}}_k(t-1),\,\varepsilon\Big), & t > t_0,
    \end{cases}
    \label{eq:penalty_ema}
\end{equation}
where $t_0$ denotes the first training step at which term $k$ is observed,
$\beta=0.1$ is the EMA momentum, and $\varepsilon=10^{-8}$ is a numerical
floor that prevents division by zero when a penalty term is exactly
satisfied (e.g., no voltage-bound violation in a batch). The
scale-normalized penalty terms
\begin{equation}
    \widetilde{\mathcal{L}}_k(t) = \frac{\mathcal{L}_k(t)}{\bar{\mathcal{L}}_k(t)},
    \qquad k \in \{V,\theta,S\},
    \label{eq:penalty_normalized}
\end{equation}
are then used to define the total loss function
\begin{equation}
    \mathcal{L}_{\mathrm{total}}
    =
    \mathcal{L}_{\mathrm{sup}}
    +
    \lambda_V \widetilde{\mathcal{L}}_{V}
    +
    \lambda_{\theta} \widetilde{\mathcal{L}}_{\theta}
    +
    \lambda_S \widetilde{\mathcal{L}}_{S}.
    \label{eq:preliminary_physics_loss}
\end{equation}
The
coefficients $\lambda_V$, $\lambda_{\theta}$, and $\lambda_S$ control the
relative contribution of the three penalties.

\subsection{Curriculum-Scheduled Penalty Method}
\label{subsec:curriculum_scheduled}
Directly imposing the full feasibility penalties from the first training
step is problematic: early in training, predictions are far from feasible,
so the penalty gradients are large and can dominate or conflict with the
supervised signal before the model has learned a reasonable approximation
of the OPF solution map. Following common practice in physics-informed and
constrained learning \cite{bengio2009curriculum, wang2021survey, krishnapriyan2021characterizing, wang2021understanding}, we instead use a curriculum schedule that keeps the
feasibility penalties inactive while the model first fits the supervised
targets, then gradually ramps them in once training has stabilized. Each
coefficient in Equation~\eqref{eq:preliminary_physics_loss} is scaled by an
epoch-dependent curriculum factor $\alpha(t)\in[0,1]$:
\begin{equation}
    \alpha(t) =
    \begin{cases}
        0, & t < t_{\mathrm{warmup}} \\
        \dfrac{t - t_{\mathrm{warmup}}}{t_{\mathrm{ramp}}}, & t_{\mathrm{warmup}} \le t < t_{\mathrm{warmup}}+t_{\mathrm{ramp}} \\
        1, & t \ge t_{\mathrm{warmup}}+t_{\mathrm{ramp}}
    \end{cases}.
    \label{eq:curriculum_schedule}
\end{equation}
\end{revision}

\section{Open-Source OPF Dataset}
\label{sec:opf_dataset}
% ============================================================

The dataset used in this work is the open-source PyTorch Geometric OPFDataset~\cite{lovett2024opfdata}, which provides heterogeneous graph instances of solved AC-OPF problems derived from Power Grid Library for Benchmarking AC Optimal Power Flow Algorithms (PGLib-OPF) cases~\cite{pglibopf}. Each sample contains grid topology, component features, operating context, and OPF solution labels. The full dataset used for model selection contains ten PGLib-OPF cases spanning more than three orders of magnitude in grid size, from 14 to 13,659 buses: case14, case30, case57, and case118 from the IEEE family; case500, case2000, and case10000 from the ARPA-E Grid Optimization Competition (GOC) family; case4661 from the Sustainable Data Evolution Technology (SDET) family; case6470 from the Réseau de Transport d’Électricité (RTE) family; and case13659 from the Pan-European Grid Advanced Simulation and State Estimation (PEGASE) family, using the original PGLib-OPF case-family labels. Each case contributes 300,000 independently drawn AC-OPF solutions, yielding a corpus of 3,000,000 heterogeneous graph instances in total. Using multiple cases is essential for the \ac{GFM} objective: a model trained only on a single topology can learn topology-specific correlations, whereas training over many cases requires the model to learn reusable representations of grid structure, operating context, and component interactions.

% ============================================================
% BEGIN COMMENTED SECTION: Scalable Data Preprocessing and HDF5 Conversion
\iffalse
\section{Scalable Data Preprocessing and HDF5 Conversion}
\label{sec:scalable_preprocessing}
% ============================================================

The raw OPF data is converted into a format suitable for distributed training by selecting cases and splits, loading raw samples, constructing typed heterogeneous graph objects with node features, edge attributes, and bus-level OPF targets, attaching graph-level context, validating feature dimensions, and serializing the processed graphs into HDF5 shards.

The full OPF corpus contains approximately three million heterogeneous graph instances. Directly loading raw JavaScript Object Notation (JSON) files during every training run would impose significant input/output (I/O) and preprocessing overhead. We therefore separate preprocessing from model training. Raw OPF samples are converted into typed heterogeneous graph objects and serialized into HydraGNN's HDF5 format. The full dataset is stored using 129 HDF5 shards, enabling scalable data access during distributed training and HPO. HDF5 conversion is essential for scalable HPO because each trial reuses the same processed dataset, avoiding repeated raw-data parsing, reducing filesystem pressure, and ensuring that all architecture comparisons use identical data.

\fi
% END COMMENTED SECTION: Scalable Data Preprocessing and HDF5 Conversion

% ============================================================
\section{HydraGNN Framework for Scalable Heterogeneous OPF Learning}
\label{sec:hydragnn_features}
% ============================================================

\subsection{Extension from Homogeneous to Heterogeneous Graphs}
\label{subsec:hydragnn_heterogeneous_extension}

HydraGNN provides scalable graph-learning infrastructure, including distributed training, configurable architectures, multi-head prediction, and high-performance data handling. The heterogeneous OPF branch extends HydraGNN from homogeneous graphs to heterogeneous graphs while preserving complete compatibility with existing homogeneous models.

The central abstraction is HeteroBase, which provides common infrastructure for all heterogeneous architectures. HeteroBase maintains separate input projectors for each node type, constructs relation-specific message-passing layers, supports variable edge-attribute dimensions, and decodes outputs on selected target node types.

\begin{revision}
Figure~\ref{fig:hydragnn_architecture} distinguishes three implementation aspects: class inheritance, factory-based model selection, and reuse of homogeneous message-passing implementations by the heterogeneous message-passing neural network (MPNN) wrappers. \texttt{HeteroBase} extends the legacy \texttt{Base} infrastructure with typed node/edge handling, heterogeneous pooling, metadata and per-type input dimensions, and shared decoding logic. Its direct subclasses expose aggregation-, attention-, relational-attention-, transformer-, edge-aware-, and PNA-based alternatives through the common \texttt{create\_model\_config}/\texttt{create\_model} factory.

\begin{figure*}[t]
    \centering
    \includegraphics[width=0.98\textwidth]{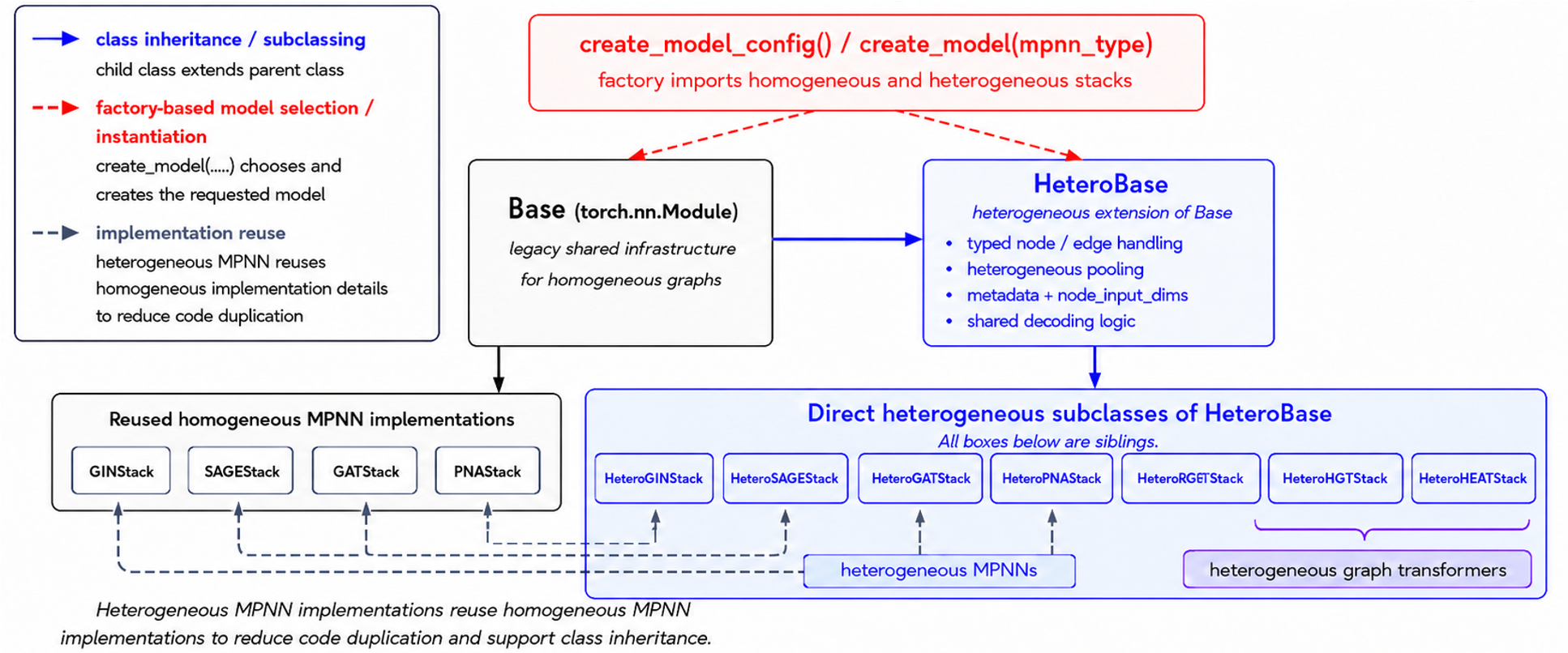}
    \caption{HydraGNN software restructuring for configurable homogeneous and heterogeneous graph models. Blue arrows denote inheritance, red dashed arrows denote factory-based selection, and gray dashed arrows denote reuse of homogeneous message-passing implementations.}
    \label{fig:hydragnn_architecture}
\end{figure*}
\end{revision}

\subsection{Unique Features of HeteroBase}
\label{subsec:heterobase_features}

The heterogeneous extension introduces several capabilities important for OPF learning:
\begin{itemize}
    \item \textbf{Per-node-type input embeddings:} separate linear projectors map bus, generator, load, and shunt features into a shared hidden space.
    \item \textbf{Per-edge-type message passing:} relation-specific convolution operators preserve the semantics of AC lines, transformers, and structural equipment links.
    \item \textbf{Variable edge-attribute dimensions:} physical edge types may carry different feature widths, e.g., nine for AC lines and eleven for transformers.
    \item \textbf{Featureless structural edges:} equipment-to-bus links can participate in message passing without continuous edge attributes.
    \item \textbf{Node-type-specific prediction:} all node types contribute to message passing, but the primary decoder is applied only to bus nodes.
    \item \textbf{Graph-level conditioning:} optional global context features can be injected through feature-wise linear modulation (FiLM), node-level concatenation, or pooled fusion.
    \item \textbf{Multi-head decoding:} the framework supports future multi-task OPF learning with graph-level and node-level outputs.
    \item \textbf{Distributed-training compatibility:} eager metadata-based initialization ensures parameters are registered before distributed data parallel wrapping.
\end{itemize}

\subsection{Supported Heterogeneous Architectures}
\label{subsec:supported_heterogeneous_architectures}

The workflow supports a family of heterogeneous architectures summarized in Table~\ref{tab:hetero_architectures}. This design treats the architecture as a fully tunable experimental variable rather than a fixed modeling assumption.
The supported families build on GraphSAGE, PNA, GAT, RGAT, HGT, and HEAT.

\begin{table*}[t]
\centering
\caption{Heterogeneous architectures supported by the HydraGNN OPF workflow.}
\label{tab:hetero_architectures}
\renewcommand{\arraystretch}{1.15}
\begin{tabular}{llll}
\toprule
\textbf{Model} & \textbf{Layer Type} & \textbf{Uses Edge Attributes?} & \textbf{Key Feature} \\
\midrule
HeteroSAGE & SAGEConv per relation & No & Stable mean aggregation with type separation \\
HeteroGAT  & GATv2Conv per relation & Yes & Dynamic attention per relation \\
HeteroRGAT & GATConv per relation & Yes & Relation-specific graph attention \\
HeteroHGT  & HGTConv & No & Type-specific key/query/value projections \\
HeteroHEAT & HEATConv & Yes & Type-aware and edge-attribute-aware attention \\
HeteroPNA  & PNAConv with bipartite wrapper & Yes & Multi-aggregator degree-scaled message passing \\
\bottomrule
\end{tabular}
\end{table*}

HeteroSAGE builds on GraphSAGE mean aggregation~\cite{hamilton2017graphsage}. HeteroGAT and HeteroRGAT build on graph-attention operators, using GATv2-style dynamic attention for HeteroGAT~\cite{brody2022gatv2} and the original GAT-style attention mechanism for HeteroRGAT~\cite{velickovic2018gat}. HeteroHGT uses the HGT operator, which introduces type-specific key, query, and value projections and relation-aware attention~\cite{hu2020hgt}. HeteroHEAT uses the HEAT operator, which combines node-type information, edge-type embeddings, and continuous edge attributes~\cite{mo2021heat}. HeteroPNA extends PNA to heterogeneous bipartite relations through a wrapper that adapts PNAConv to source-target type pairs. In all cases, the HydraGNN contribution is not the algorithmic design of these base layers, but their integration into a common distributed heterogeneous OPF workflow with type-specific encoders, relation-specific message passing, scalable data loading, HPO, and downstream fine-tuning support.

\subsection{Layer Provenance and HydraGNN New Contributions}
\label{subsec:layer_provenance}

The heterogeneous OPF models combine established graph-learning layers with new HydraGNN infrastructure. The base message-passing primitives are drawn from the literature: GraphSAGE for inductive neighborhood aggregation, GAT and GATv2 for attention-based message passing, PNA for multi-aggregator degree-scaled message passing, HGT for type-aware graph-transformer attention, and HEAT for edge-attribute-aware heterogeneous attention. In addition to the features mentioned in section \ref{subsec:hydragnn_heterogeneous_extension}, HydraGNN provides node-type-specific feature projections, relation-specific layer construction, support for variable edge-attribute dimensions, bus-level target decoding, OPF-domain regularization, HDF5-based distributed data loading, and Frontier-scale HPO across all candidate architectures.

% ============================================================
\section{Scalable Hyperparameter Optimization}
\label{sec:scalable_hpo}
% ============================================================

\subsection{DeepHyper-Based HPO}
\label{subsec:deephyper_hpo}

To identify architectures and hyperparameters suitable for full-dataset OPF surrogate modeling, we performed HPO using DeepHyper~\cite{balaprakash2018deephyper,egele2025deephyper}. Its centralized Bayesian optimization strategy maintains a central surrogate model while parallel workers evaluate candidate configurations. The entire data corpus was used for the training of each GNN architecture selected by the HPO. We used an Upper Confidence Bound acquisition function, which balances exploitation of configurations with favorable predicted performance against exploration of configurations with high predictive uncertainty~\cite{snoek2012practical,srinivas2010gpucb}. The search space is summarized in Table~\ref{tab:hpo_search_space}.

\begin{table}[t]
\centering
\caption{Hyperparameter search space for full-dataset HPO.}
\label{tab:hpo_search_space}
\renewcommand{\arraystretch}{1.15}
\begin{tabular}{lll}
\toprule
\textbf{Hyperparameter} & \textbf{Range} & \textbf{Type} \\
\midrule
hidden\_dim & $[32,256]$ & Integer \\
num\_conv\_layers & $[2,6]$ & Integer \\
learning\_rate & $[10^{-5},10^{-2}]$ & Log-uniform \\
mpnn\_type & fixed per HPO run & Categorical \\
\bottomrule
\end{tabular}
\end{table}

We launched six architecture-specific HPO runs, one per GNN architecture (HeteroSAGE, HeteroHEAT, HeteroHGT, HeteroPNA, HeteroRGAT, and HeteroGAT). Fixing the architecture within each HPO run allows the optimizer to search the continuous and integer hyperparameters under the same compute budget and data distribution.

\subsection{Frontier Execution Strategy}
\label{subsec:frontier_hpo_execution}

\begin{revision}
The entire campaign was executed on the supercomputer Frontier hosted by the Oak Ridge Leadership Computing Facility \cite{olcffrontier}. Each architecture-specific campaign allocated 512 nodes (4,096 AMD MI250X GPUs). Each HPO trial used 32 nodes (256 GPUs), allowing as many as 16 concurrent trials. Every HPO trial completed ten training epochs with batch size 32 and processed the entire dataset in each epoch. The ten-epoch budget was an optimization budget, not a common wall-clock cutoff, thus allowing each model to run as long as needed to complete the 10 epochs. Failed configurations were recorded for execution-robustness analysis but excluded from validation-loss distributions and predictive ranking.

The objective is the minimum validation loss among the ten completed epochs. This choice accommodates oscillatory validation trajectories, but it does not ensure that every architecture reached asymptotic convergence. In fact, attention- and transformer-based models may require longer schedules, so the reported ranking is explicitly conditional on the common ten-epoch optimization budget. 
\end{revision}

\subsection{Fault Tolerance and Result Reuse}
\label{subsec:hpo_fault_tolerance}
\begin{revision}
At this scale, a trial may fail when the sampled model width and depth, the memory required by its relation-specific layers, the distributed-training configuration exceed the capabilities of the allocated GPUs, or the new HPO trials is simply selected and started too late and cannot complete within the hard bounds of the job scheduled on the supercomputing facility. The framework isolates trial failures so that one failure does not interrupt the campaign. Only configurations that completed the prescribed ten epochs and produced a valid objective enter the predictive comparison; completion counts are reported separately as workflow-level execution information. Results are stored in a reusable comma-separated-value (CSV) representation that can initialize follow-up searches.
\end{revision}

\section{Fine-Tuning Protocol}
\label{subsec:ft_setup}

\begin{revision}
After pretraining, the bus-node prediction head
$f_{\Theta_{\mathrm{head}}}^{\mathrm{bus}}$ in
Equation~\eqref{eq:bus_prediction_head} is replaced with a
task-specific prediction head. 
Our downstream experiments compare scratch training, full fine-tuning, partial fine-tuning, and head-only fine-tuning as a function of labeled training volume. A detailed description of what parameter set is replaced, frozen, or adapted for each strategy is described in Table \ref{tab:adaptation_regimes}.
The same training subsets and validation partition are used for every adaptation regime. All downstream runs use learning rate $10^{-4}$ and 50 epochs on 8 Frontier nodes (64 AMD MI250X GPUs).
\begin{table*}[t]
\centering
\scriptsize
\setlength{\tabcolsep}{4pt}
\renewcommand{\arraystretch}{1.12}
\caption{Initialization and trainable components in the four downstream regimes. ``New'' denotes a task-specific head initialized for FT1 or FT2.}
\label{tab:adaptation_regimes}
\begin{tabular}{lccccc}
\toprule
\textbf{Regime} & \textbf{Initialization} & \textbf{Input projectors} & \textbf{Message-passing layers} & \textbf{Final MP layer} & \textbf{Prediction head} \\
\midrule
Scratch & Random & Trainable & Trainable & Trainable & Trainable \\
Full fine-tuning & Pretrained & Trainable & Trainable & Trainable & New/trainable \\
Partial fine-tuning & Pretrained & Frozen & Frozen except final & Trainable & New/trainable \\
Head-only fine-tuning & Pretrained & Frozen & Frozen & Frozen & New/trainable \\
\bottomrule
\end{tabular}
\end{table*}
\end{revision}

% ============================================================
\section{Results}
\label{sec:results_and_gfm_protocol}
% ============================================================

\begin{revision}
\subsection{Supervised-Only Training vs. Feasibility-Aware Penalties}
\label{subsec:feasibility_ablation_results}

To empirically assess the preliminary objective of
Section~\ref{subsec:preliminary_feasibility_losses}, we compared two HeteroHEAT models with the same
hyperparamater configuration (hidden dimension 250, 6 message-passing layers, 4 attention heads,
learning rate $10^{-2}$): one trained
with the supervised loss $\mathcal{L}_{\mathrm{sup}}$ alone, and one trained
with the full feasibility-aware objective $\mathcal{L}_{\mathrm{total}}$ of
Equation~\eqref{eq:preliminary_physics_loss} using
$\lambda_V=\lambda_{\theta}=\lambda_S=0.01$. Both
models were trained for the 10 epochs. The curriculum-scheduled penalty method used $t_{\mathrm{warmup}}=3$ and $t_{\mathrm{ramp}}=3$, so the feasibility
penalties reaches their full configured weight only at epoch 6 of the
10-epoch training run.

Table~\ref{tab:feasibility_ablation} reports the supervised validation
MSE for both configurations, together with the individual
feasibility-penalty terms evaluated on the validation set. 
\begin{table}[t]
    \centering
    \caption{Ablation comparing supervised-only training against the
    feasibility-aware objective of
    Equation~\eqref{eq:preliminary_physics_loss} for the HeteroHEAT
    architecture. Values are computed on the held-out validation split.}
    \label{tab:feasibility_ablation}
    \begin{tabular}{lcc}
        \toprule
        Metric & Supervised-only & Feasibility-aware \\
        \midrule
        Validation MSE ($\mathcal{L}_{\mathrm{sup}}$) & 0.03701 & 0.08807 \\
        Voltage-bound penalty ($\mathcal{L}_{V}$)      & $1.25\times10^{-5}$ & $2.50\times10^{-6}$ \\
        Angle-difference penalty ($\mathcal{L}_{\theta}$) & $4.48\times10^{-3}$ & $<10^{-8}$ \\
        Thermal-limit penalty ($\mathcal{L}_{S}$)       & 377.80 & 0.360 \\
        \bottomrule
    \end{tabular}
\end{table}
The addition of the feasibility-aware terms did not improve the supervised
validation MSE used for model selection. 

The results can be interpreted as a consequence of competing training
objectives. The supervised data-driven term rewards agreement with the
reference OPF solutions, whereas the feasibility penalties reward
satisfaction of selected operating constraints, and combining them through
a weighted sum requires the optimizer to balance objectives that can
induce conflicting gradient directions during training. This competition
is directly visible in the per-epoch training and validation validation loss shown in Figure \ref{fig:loss_curves_feasibility}: while the
feasibility penalties are held at zero weight (epochs 0--3 of the
curriculum schedule in Equation~\eqref{eq:curriculum_schedule}), the
validation loss decreases monotonically; once the penalties ramp in
(epochs 4--6), it jumps sharply and then plateaus at a substantially
higher value for the remainder of training, indicating that satisfying the
feasibility constraints pulls the optimizer away from the region that
minimizes the supervised objective alone. 
\begin{figure}[t]
  \centering
  \includegraphics[width=\linewidth]{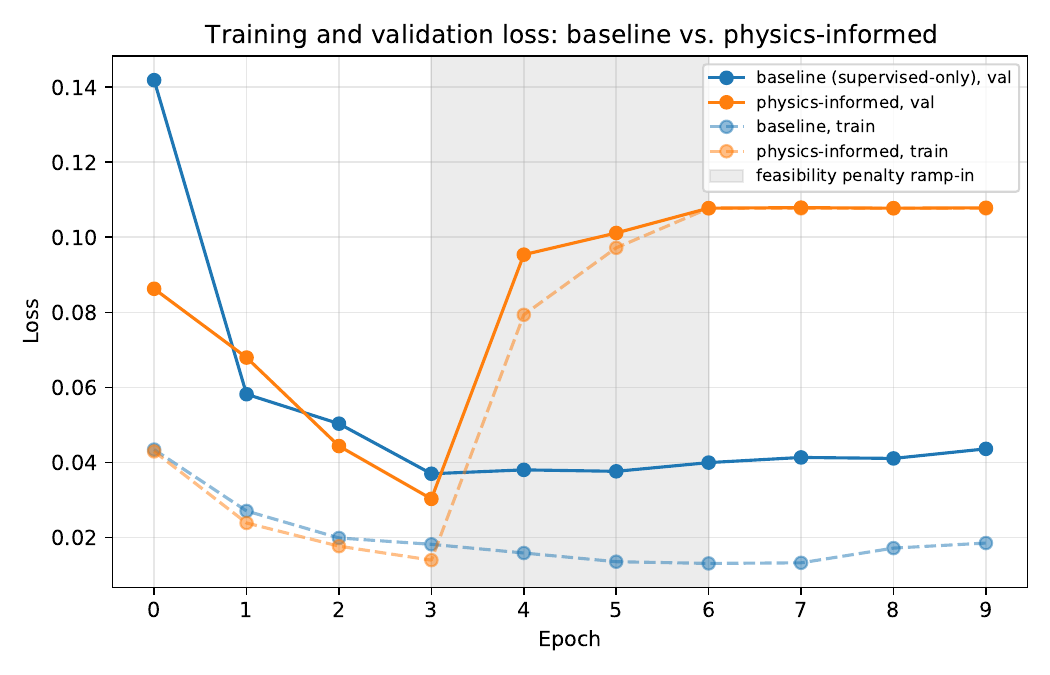}
  \caption{Per-epoch training (dashed) and validation (solid) loss for the
  supervised-only baseline and the physics-informed (feasibility-penalty)
  training runs on the same HeteroHEAT configuration. The shaded region
  marks the curriculum ramp-in of the feasibility penalty weights
  (epochs 3--6, Equation~\eqref{eq:curriculum_schedule}).}
  \label{fig:loss_curves_feasibility}
\end{figure}
This outcome motivated our decision
to use $\mathcal{L}_{\mathrm{sup}}$ alone for the full-corpus HPO campaign, in
order to maintain a consistent and computationally tractable comparison
across the six heterogeneous architecture families.

{\color{black}We also note that supervised MSE and operating-limit penalty values measure different aspects of the predictions and neither is, by itself, a comprehensive evaluation of physical feasibility.} Such an evaluation would additionally
consider voltage-magnitude and voltage-angle errors in physical units, nodal
active and reactive power balance residuals, generator limit violations,
branch thermal limit violations, voltage bound and angle difference
violations, and, where appropriate, solver recovery or warm-start
performance.

\begingroup\color{black}
\subsection{Zero-Shot Evaluation on an Unseen Grid Topology}
\label{subsec:rts73_zero_shot}

To examine whether the observed accuracy--constraint trade-off extends beyond the
pretraining systems, we evaluated the same supervised-only and
feasibility-aware HeteroHEAT models on the previously unseen
\texttt{pglib\_opf\_case73\_ieee\_rts} network. This 73-bus topology was
not among the ten systems used for pretraining or HPO. The experiment
therefore evaluates out-of-topology prediction without retraining or
fine-tuning. Both models were evaluated under the same protocol, using
the supervised squared-error objective and the three physical penalties
already defined in Section~\ref{subsec:preliminary_feasibility_losses}.

Table~\ref{tab:rts73_zero_shot} summarizes the results. Relative to the
supervised-only model, feasibility-aware training increased the test MSE, while significantly reducing the voltage-bound penalty, the angle-difference penalty, and the thermal-limit. Thus, the trade-off seen on the original
validation data persists on this unseen topology: the physics-informed
objective produces lower values of all three evaluated operating-limit
penalties, despite a higher supervised prediction error.

\begin{table}[t]
\color{black}
\centering
\caption{Zero-shot evaluation of the matched HeteroHEAT models on the
unseen IEEE RTS 73-bus topology. The MSE and physical penalty terms
are evaluated on the new topology.}
\label{tab:rts73_zero_shot}
\begin{tabular}{lcc}
\toprule
Metric & Supervised-only & Feasibility-aware \\
\midrule
Test MSE ($\mathcal{L}_{\mathrm{sup}}$) & 0.0107 & 0.0303 \\
Voltage-bound penalty ($\mathcal{L}_{V}$) & $3.23\times10^{-5}$ & $1.21\times10^{-6}$ \\
Angle-difference penalty ($\mathcal{L}_{\theta}$) & $0.0298$ & $4.35\times10^{-6}$ \\
Thermal-limit penalty ($\mathcal{L}_{S}$) & 9.2811 & 0.0143 \\
\bottomrule
\end{tabular}
\end{table}
\endgroup

\subsection{Full-Dataset HPO Results}
\label{subsec:full_dataset_hpo_results}

The full HPO campaign dispatched 387 trials across six architectures and successfully completed 153 of them. Figure~\ref{fig:full_dataset_hpo_boxplot} visualizes the trial-level distributions per architecture, with valid/dispatched counts on the x-axis and the best trial marked. 

\begin{revision}
Table~\ref{tab:hpo_summary} provides the numerical campaign summary. Medians and interquartile ranges are read from the retained trial distribution and are reported to the precision supported by Figure~\ref{fig:full_dataset_hpo_boxplot}; HeteroPNA is excluded from distributional statistics because only two of 115 trials completed.

\begin{table*}[t]
\centering
\scriptsize
\setlength{\tabcolsep}{5pt}
\renewcommand{\arraystretch}{1.12}
\caption{Full-corpus HPO summary under the common ten-epoch protocol. Approximate medians and interquartile ranges (IQRs) summarize the plotted completed-trial distributions.}
\label{tab:hpo_summary}
\begin{tabular}{lcccccp{3.5cm}}
\toprule
\textbf{Architecture} & \textbf{Valid / dispatched} & \textbf{Completion} & \textbf{Best MSE} & \textbf{Median MSE} & \textbf{IQR} & \textbf{Interpretation} \\
\midrule
HeteroSAGE & 16/32 & 50.0\% & 0.0166 & $\sim$0.0336 & $\sim$[0.0257,0.0389] & Lowest loss and comparatively narrow spread \\
HeteroHEAT & 16/32 & 50.0\% & 0.0167 & $\sim$0.0338 & $\sim$[0.0284,0.0453] & Comparable best loss; greater HPO sensitivity \\
HeteroRGAT & 73/97 & 75.3\% & 0.0333 & $\sim$0.0450 & $\sim$[0.0395,0.0564] & Highest completion rate; second predictive tier \\
HeteroHGT & 10/48 & 20.8\% & 0.0336 & $\sim$0.0480 & $\sim$[0.0435,0.0535] & Few completed trials; ranking uncertainty is larger \\
HeteroGAT & 36/63 & 57.1\% & 0.0367 & $\sim$0.0595 & $\sim$[0.0470,0.0634] & Highest median loss among compared distributions \\
HeteroPNA & 2/115 & 1.7\% & -- & -- & -- & Reported for completeness only; no statistical comparison \\
\bottomrule
\end{tabular}
\end{table*}
\end{revision}

Figure~\ref{fig:full_dataset_hpo_boxplot} shows the per-trial validation-loss distributions as box plots with overlaid points and starred best trials, capturing both peak performance and typical behavior across the search space. Figure~\ref{fig:hpo_val_loss_vs_params} complements this view by plotting validation loss against trainable parameter count: HeteroSAGE ($\sim$1.6M params) and HeteroHEAT ($\sim$1.7M) reach the lowest losses at substantially smaller model sizes than HeteroRGAT ($\sim$3.5M), HeteroHGT ($\sim$3.2M), and HeteroGAT ($\sim$3.2M), none of which match the top tier even at much larger budgets. {\it Crucially, the HPO results reveal no evident correlation between model accuracy and model size: larger parameter counts do not systematically yield lower validation MSE, and the best-performing architectures are among the most compact.}

{\textcolor{black}{
Importantly, the identification of compact, high-performing models is itself an outcome of the computationally intensive HPO campaign. Systematically exploring diverse architectures and hyperparameter configurations, each trained on millions of heterogeneous OPF graphs, requires substantial computational resources. Leadership-class computing facilities make such extensive exploration practical through distributed training and concurrent evaluations. Thus, the modest parameter counts of the selected models do not diminish the importance of large-scale computing; rather, they demonstrate its role in enabling the discovery of accurate, parameter-efficient heterogeneous GNN architectures.}
}

\textbf{Leading architectures: HeteroSAGE and HeteroHEAT.}
HeteroSAGE and HeteroHEAT achieve essentially identical best validation MSE, suggesting both have reached a similar accuracy ceiling within the 10-epoch HPO budget. Both also have the lowest median validation MSE, with only a small gap between them (Figure~\ref{fig:full_dataset_hpo_boxplot}), confirming that their superiority is not a single-trial artifact.

\textbf{Sensitivity to hyperparameters.}
Despite comparable peak and median performance, HeteroSAGE and HeteroHEAT differ in hyperparameter sensitivity, as measured by the cross-trial spread (Figure~\ref{fig:full_dataset_hpo_boxplot}). HeteroHEAT has a noticeably wider distribution, indicating greater sensitivity to hidden dimension, depth, and learning rate. This matches the architectural difference: simple mean aggregation in HeteroSAGE is inherently stable, while the type- and edge-aware attention in HeteroHEAT can amplify or suppress information depending on the learning dynamics. For deployment, HeteroSAGE is the more robust default; HeteroHEAT reaches comparable accuracy but needs more careful tuning.

\textbf{Second-tier architectures.}
HeteroRGAT (73 valid trials) and HeteroHGT (10 valid trials) form a second tier, with best validation MSE roughly twice those of the leaders (Figure~\ref{fig:full_dataset_hpo_boxplot}). They are essentially tied on best loss but differ in spread: HeteroHGT is more concentrated around its mean, though its low completion rate limits confidence in the comparison and suggests that many configurations exhaust resources before producing reliable results. HeteroGAT performs worst among the architectures with a usable trial set; its median exceeds its mean, so most trials underperform the average and the best value is an outlier.

\textbf{HeteroPNA stability issues.}
HeteroPNA produced only two valid trials out of 115 dispatched (under 2\% completion), attributable to the bipartite PNA wrapper used to extend PNAConv to heterogeneous graphs, which imposes higher memory and communication overhead at 256 GPUs per trial. The two surviving trials nominally place HeteroPNA between the top and second tiers, but reflect an extreme selection effect---only the easiest (likely smallest hidden dimension) configurations succeeded---so no reliable mean, standard deviation, or median can be reported.

Because the architecture comparison uses a common ten-epoch training budget,
HPO with a longer training budget per trial could alter the relative ranking
of the evaluated architectures.

\begin{figure}[t]
    \centering
    \includegraphics[width=\columnwidth]{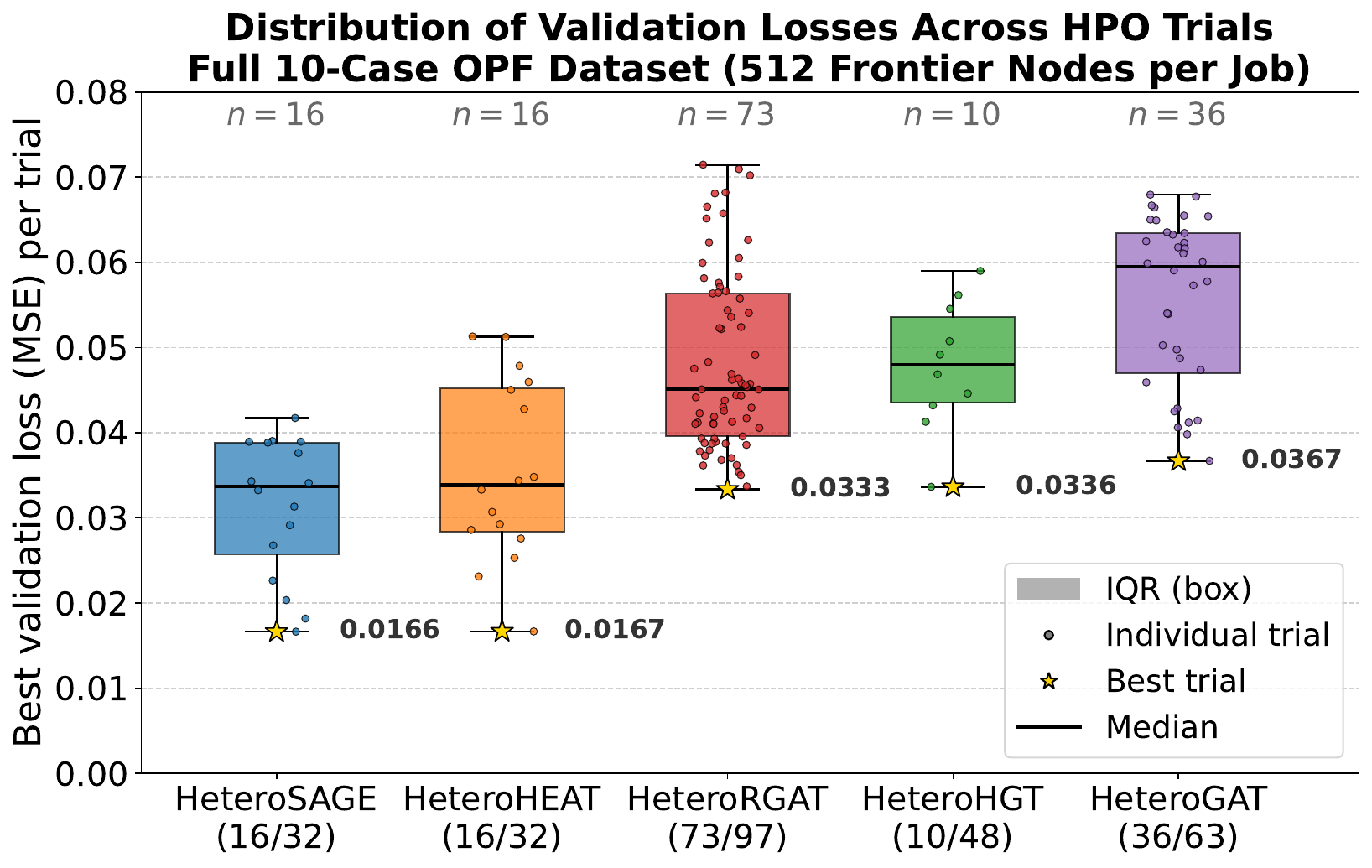}
    \caption{Distribution of per-trial best validation losses across HPO trials for each GNN architecture. X-axis labels report architecture name and valid/dispatched trial counts. Boxes span the 25th--75th percentile; whiskers extend to 1.5$\times$ the interquartile range; jittered points overlay individual trials; gold stars mark the best trial per architecture. The value $n$ denotes the number of successful HPO trials for each GNN architecture.}
    \label{fig:full_dataset_hpo_boxplot}
\end{figure}

\begin{figure}[t]
    \centering
    \includegraphics[width=\columnwidth]{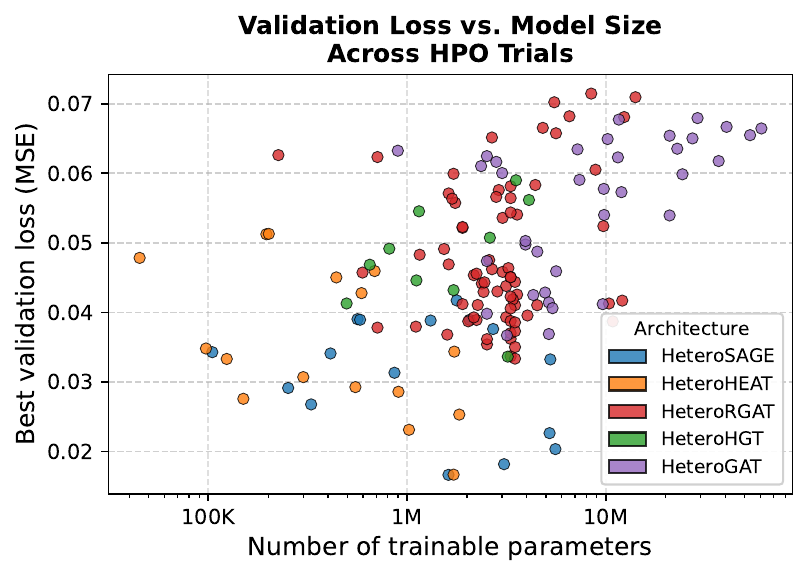}
    \caption{Validation loss vs.\ number of trainable parameters per HPO trial on a log scale, colored by architecture.}
    \label{fig:hpo_val_loss_vs_params}
\end{figure}

\subsection{Model Selection for Downstream Experiments}
\label{subsec:model_selection}

Based on the full-dataset HPO campaign, we select HeteroSAGE and HeteroHEAT as the primary architectures for subsequent pretraining and fine-tuning. HeteroSAGE is the default scalable baseline: it attains the best validation MSE loss with the lowest cross-trial spread (Figure~\ref{fig:full_dataset_hpo_boxplot}), so its advantage is robust to hyperparameter variation rather than tied to a narrow region of the search space. HeteroHEAT is the edge-aware attention alternative, with comparable peak accuracy and median while explicitly leveraging continuous edge attributes and type embeddings. HeteroRGAT and HeteroHGT are retained as secondary baselines, since their distinct architectural families (relational attention and type-aware transformer) may behave differently under topology-specific fine-tuning. HeteroGAT shows consistently higher losses and an unfavorable median-to-mean ratio, indicating limited suitability for the full-scale task.

\subsection{Strong and Weak Scaling Results}
\label{subsec:full_dataset_scaling_results}
The scaling experiments were performed on Frontier using eight AMD MI250X GPUs per node and the OPF data corpus containing approximately 2.7 million training heterogeneous graphs. The two HeteroSAGE and HeteroHEAT models selected by HPO were used for the scaling tests, each trained for three epochs. The reported scaling times correspond to the execution of HydraGNN’s \texttt{train\_validate\_test} routine, while data-loading overhead was measured and reported separately. For strong scaling, the global batch size was fixed at 32,768 graphs and each epoch was limited to 20 batches, corresponding to 1,966,080 graph presentations per run. 

The HeteroSAGE measurements span 16--1,024 nodes, for which the local batch decreases from 256 to 4 graphs per GPU. HeteroHEAT requires more device memory: attempted 16-, 32-, and 64-node runs exceeded the 64~GiB memory available per GPU at their required local batches, so its valid fixed-workload series begins at 128 nodes. For weak scaling, the local batch was fixed at 32 graphs per GPU and the global batch increased from 256 graphs on one node to 262,144 graphs on 1,024 nodes. Ten batches per epoch were used so that the largest-scale run did not reuse samples within an epoch.

Figure~\ref{fig:strong_scaling_efficiency} reports strong scaling parallel efficiency relative to the smallest successful node count for each model. HeteroSAGE is nearly ideal from 16 to 32 nodes: execution time decreases from 225.34 to 113.87~s, corresponding to 98.9\% efficiency. Time continues to decrease overall through 256 nodes, where the minimum of 50.96~s is observed, before increasing to 71.94 and 125.15~s at 512 and 1,024 nodes. HeteroHEAT decreases from 133.79~s at its 128-node baseline to a minimum of 117.44~s at 256 nodes, then increases to 137.47 and 166.41~s. At the largest scales, communication and synchronization dominate the progressively smaller computation assigned to each GPU.

\begin{figure}[t]
    \centering
    \includegraphics[width=\columnwidth]{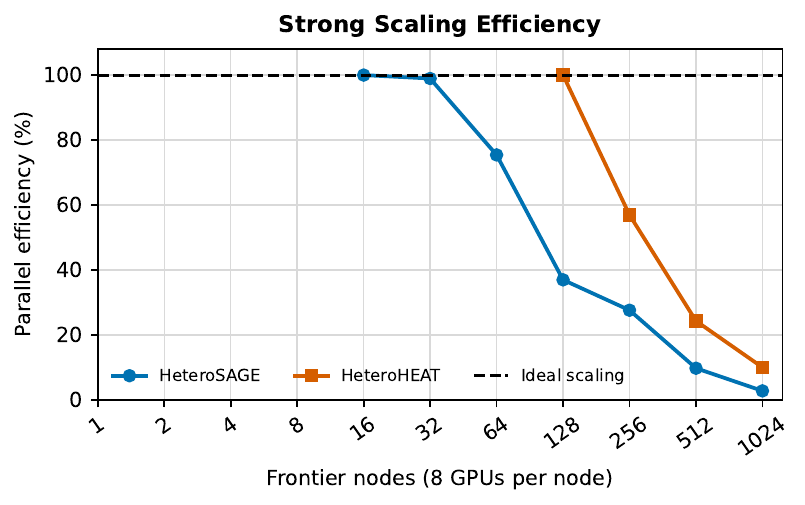}
    \caption{Strong scaling efficiency for a fixed global batch of 32,768 graphs and 20 batches per epoch. Efficiency is normalized independently to the smallest successful run for each architecture (16 nodes for HeteroSAGE and 128 nodes for HeteroHEAT). The dashed black line denotes ideal 100\% efficiency.}
    \label{fig:strong_scaling_efficiency}
\end{figure}

Figure~\ref{fig:weak_scaling_efficiency} reports weak scaling efficiency relative to each model's one-node execution time. At 256 nodes, HeteroSAGE and HeteroHEAT retain 61.6\% and 80.1\% efficiency, respectively. Efficiency decreases more sharply at 512 and 1,024 nodes, reaching 26.2\% for HeteroSAGE and 46.5\% for HeteroHEAT at 1,024 nodes. Although weak scaling efficiency decreases beyond 256 nodes, aggregate throughput continues to increase, reaching approximately 84,372 graphs/s for HeteroSAGE and 57,379 graphs/s for HeteroHEAT at 1,024 nodes. Together, the strong and weak scaling results identify approximately 256 nodes as the practical efficiency limit for this measured workload. They also show that HeteroHEAT has a larger memory footprint and a higher absolute execution cost than HeteroSAGE.

\begin{figure}[t]
    \centering
    \includegraphics[width=\columnwidth]{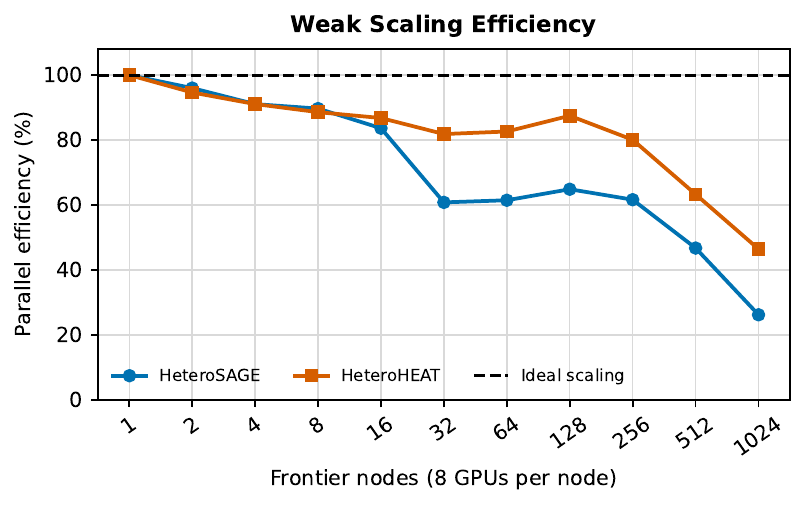}
    \caption{Weak scaling efficiency with a fixed local batch of 32 graphs per GPU and 10 batches per epoch. The global batch increases from 256 graphs on one node to 262,144 graphs on 1,024 nodes. The dashed black line denotes ideal 100\% efficiency.}
    \label{fig:weak_scaling_efficiency}
\end{figure}
\end{revision}

\subsection{FT1: Extreme-Overload Classification}
\label{subsec:ft1_results}

\begin{figure*}[t]
    \centering
    \includegraphics[width=0.95\textwidth]{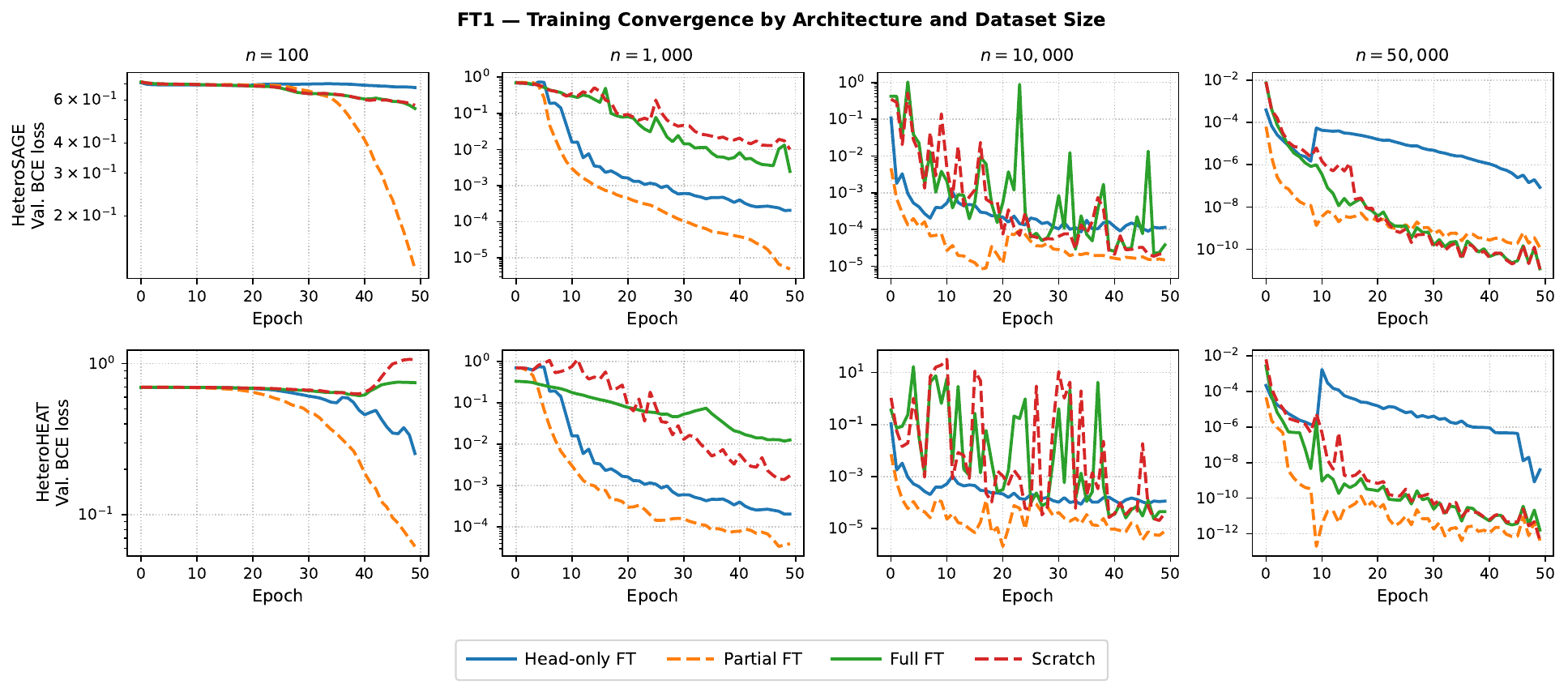}
    \caption{FT1 feasibility classification: validation BCE loss vs.\
    training epoch for HeteroSAGE (top row) and HeteroHEAT (bottom row) at four
    sample sizes. }
    \label{fig:ft1_convergence}
\end{figure*}

\begin{revision}
\textbf{Fine-tuning task 1 (FT1)} is a binary classification task to distinguish nominal operating conditions from artificially generated extreme-overload conditions. Positives are feasible case118 samples from OPFDataset~\cite{lovett2024opfdata}; negatives are generated by multiplying every active and reactive load by six. This construction separates the two classes by a deliberately extreme aggregate-demand change.

The final node embeddings of the models are pooled into a graph-level representation and passed to a binary-classification head. 
The experiment tests whether the pretrained encoder and parameter-freezing regimes can be connected to a graph-level binary head trained with binary cross-entropy (BCE), and it operates as a sanity check on the performance of the GFMs.
Most architecture--regime combinations reach near-perfect validation accuracy with $n=100$ labeled training samples. 
Figure~\ref{fig:ft1_convergence} reports validation BCE in addition to classification accuracy. The consistently low BCE complements the near-perfect accuracy results and confirms effective separation between the nominal and extreme-overload operating conditions. The results show that the pretrained heterogeneous representations can be adapted effectively to a graph-level objective with very limited labeled data and that all evaluated parameter-freezing regimes can successfully optimize the new classification head. 
\end{revision}

\begin{figure*}[t]
    \centering
    \includegraphics[width=0.95\textwidth]
    {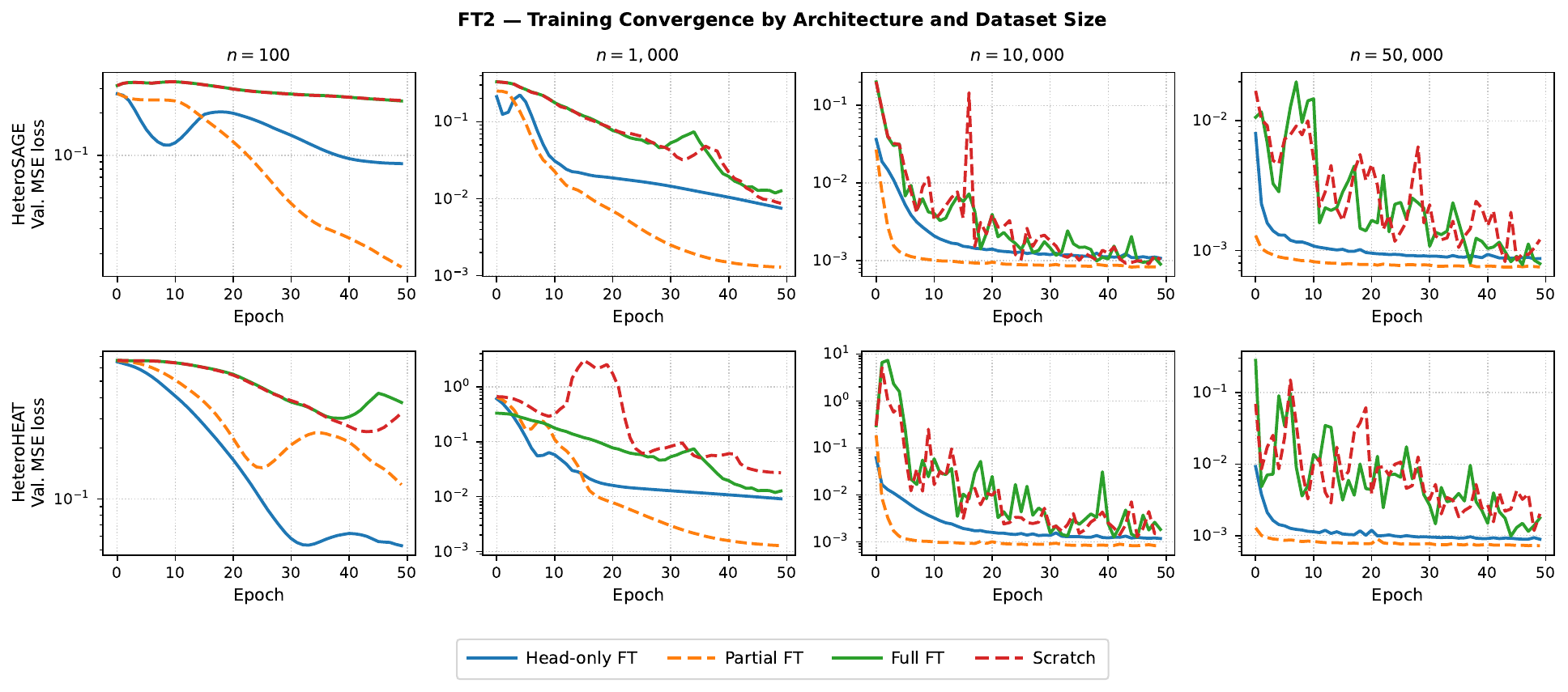}
    \caption{FT2 $N{-}1$ contingency regression: normalized
    validation MSE versus training epoch for HeteroSAGE and HeteroHEAT across
    multiple labeled-data regimes and adaptation strategies.}
    \label{fig:ft2_convergence}
\end{figure*}

\begin{revision}
\subsection{FT2: N-1 Contingency OPF Regression}
\label{subsec:ft2_results}

\textbf{Fine-tuning task 2 (FT2)} is a regression task using $N{-}1$ single-component-outage configurations on the \texttt{pglib\_opf\_case118\_ieee} grid topology. We retain all 300,000 samples across all 20 OPFData groups without filtering: generator, AC-line, and transformer outages are all included. The index-based 90/5/5 split supplied by PyTorch Geometric's OPFDataset yields 270,000 training, 15,000 validation, and 15,000 test samples. For labeled-data experiments, only the training partition is truncated to $n\in\{100,1{,}000,10{,}000,50{,}000\}$; validation and test partitions remain fixed. Splits are defined by sample index rather than outage identity, so the same outage topology may occur under different operating conditions in multiple partitions and no outage identity is explicitly held out. FT2 therefore evaluates adaptation to unseen operating samples under familiar $N{-}1$ contingencies. Target labels that need to be predicted by the fine-tuned GFMs are bus voltage angle and magnitude under the perturbed topology.

The models use a new bus-node regression head to predicts voltage angle and voltage magnitude under the perturbed topology.

Figure~\ref{fig:ft2_convergence} compares training from
scratch, full fine-tuning, partial fine-tuning, and head-only fine-tuning for
HeteroSAGE and HeteroHEAT across multiple labeled-data regimes.
Partial fine-tuning achieves lower
validation MSE than training from scratch for both HeteroSAGE and HeteroHEAT
across the evaluated low-data regimes. It also produces smoother validation
trajectories than scratch training and full fine-tuning. 
This is due to the fact that an outage involving a generator, AC line, or transformer changes the graph
topology and, consequently, the message-passing context encountered by the
model. In this situation. head-only fine-tuning preserves all pretrained message-passing layers,
which limits the model's ability to adjust its internal representation to the
perturbed topology. Partial fine-tuning addresses this limitation by updating
the prediction head and the final message-passing layer while retaining the
representations learned by the earlier layers.
These results show
that the representations learned during multi-case pretraining provide an
effective initialization for adapting the models to $N{-}1$ contingency
regression.

Figure~\ref{fig:ft2_convergence} also shows that full fine-tuning exhibits
greater validation variability in the evaluated low-data setting. Partial
fine-tuning therefore provides the strongest observed balance among adaptation
capacity, validation performance, and optimization stability. It additionally
updates fewer trainable parameters and requires less optimizer-state storage
than full fine-tuning or training from scratch.
\end{revision}

% ============================================================
\section{Conclusions}
\label{sec:conclusion}

\begin{revision}
This work advances the SOTA in graph-based AC-OPF learning by
moving the focus from individual surrogate models to an integrated and
scalable model-development workflow. Previous studies established several capabilities separately: message-passing models for power-flow and OPF
approximation, heterogeneous representations of grid components,
feasibility-aware objectives, common interfaces for comparing GNN
architectures, and multi-topology pretraining and transfer. Our work
builds on these contributions and connects them with the distributed
infrastructure required to establish a connection between grid foundation-model development
and leadership-class computing and to efficiently scale and perform model selection and adaptation over
millions of heterogeneous power system graphs.  

More specifically, the extension of HydraGNN developed in this work enabled the following five major advances.
\begin{enumerate}[label=(\roman*)]
\item Turn the selection of
heterogeneous GNN architecture into a controlled experimental variable.
Message-passing, attention-based, relational, edge-aware, degree-scaled, and
graph transformer models can be trained through the same graph
representation, data pipeline, prediction interface, and evaluation
protocol. This integration enables architectural conclusions to be drawn
under consistent conditions while preserving the physical distinctions
among grid components and relations. 
\item Perform large-scale HPO to dispatch large ensembles of GNN architectures to approximate AC-OPF solutions, and
establish a systematic approach for determining which model families are
effective and computationally robust at full-corpus scale. 
\item Assess the effect of model size over the predictive accuracy across a diverse set of GNN families, within the evaluated search space and within a prescribed optimization budget. 
\item Perform a strong and weak scaling focused on the GNN architectures identified as best performing by HPO to identify the best compromise between data throughput and scaling efficiency, which could support an efficient future re-training of the same architectures on updated (possibly larger) data corpora.
\item Integrate multiple parameter-efficient fine-tuning strategies within the
same framework to allow systematic comparisons among full, partial, head-only, and scratch training. This comparison shows
that effective adaptation depends on which portions of the pretrained model
are allowed to respond to changes in the task and graph structure.
\end{enumerate}

\section*{Data and Code Availability}

The processed datasets and pretrained GFMs produced in
this study are publicly available through the Oak Ridge Leadership Computing
Facility Data Constellation~\cite{lupopasini2026opfgfm}. The release includes
the packed HDF5 datasets used for multi-case pretraining, feasibility
classification, and $N{-}1$ contingency regression; pretrained HeteroSAGE and
HeteroHEAT model checkpoints; hyperparameter-optimization summaries; and
aggregated fine-tuning results. The release is available at
\url{https://doi.org/10.13139/OLCF/3239722}.

The HydraGNN source code with heterogeneous graph
implementations, preprocessing utilities, distributed-training workflow, and
configurations required to reproduce the experiments is publicly available at \url{https://github.com/allaffa/HydraGNN/tree/heterogeneous_graph_transformers_and_heterogeneous_gps}.

\section*{Declaration on the Use of Generative AI}

Generative AI tools were used for limited language editing and to improve
textual clarity. They were not used to generate data, conduct experiments,
produce scientific results, or determine the conclusions of this manuscript.
The authors reviewed and revised all AI-assisted text and take full
responsibility for the manuscript's content.

\end{revision}

% ============================================================
% ============================================================
% Place these near the relevant sections after final text tightening.
%
% \begin{figure*}[t]
%     \centering
%     \includegraphics[width=0.95\textwidth]{fig01_heterogeneous_opf_graph_schema.png}
%     \caption{Heterogeneous OPF graph representation used in this work.}
%     \label{fig:heterogeneous_opf_schema}
% \end{figure*}
%
% \begin{figure*}[t]
%     \centering
%     \includegraphics[width=0.95\textwidth]{fig02_scalable_opf_hydragnn_workflow.png}
%     \caption{Scalable OPF graph foundation-model workflow implemented in HydraGNN.}
%     \label{fig:opf_workflow}
% \end{figure*}

\bibliographystyle{IEEEtran}
\bibliography{references}

\end{document}